\documentclass[runningheads]{llncs}
\usepackage[T1]{fontenc}
\usepackage{graphicx}
\usepackage{booktabs}
\usepackage{amssymb}
\begin{document}
\title{Rigid Single-Slice-in-Volume registration via rotation-equivariant 2D/3D feature matching}
\titlerunning{SLIV-Reg}
\author{Anonymous}
\author{Stefan Brandst\"atter, Philipp Seeb\"ock, Christoph F\"urb\"ock, Svitlana Pochepnia, Helmut Prosch, Georg Langs}
\authorrunning{S. Brandstätter et al.}
\institute{Computational Imaging Research Lab, Department of Biomedical Imaging and Image-guided Therapy, Medical University of Vienna, Austria 
\email{xxx@meduniwien.ac.at} \url{https://www.cir.meduniwien.ac.at}
\and
Division of General Radiology, Department of Biomedical Imaging and Image-guided therapy, Medical University Vienna
\and
Christian Doppler Laboratory for Machine Learning Driven Precision Medicine, Department of Biomedical Imaging and Image-guided therapy, Medical University Vienna\
}
\maketitle              % typeset the header of the contribution
\begin{abstract}
2D to 3D registration is essential in tasks such as diagnosis, surgical navigation, environmental understanding, navigation in robotics, autonomous systems, or augmented reality. In medical imaging, the aim is often to place a 2D image in a 3D volumetric observation to w. Current approaches for rigid single slice in volume registration are limited by requirements such as pose initialization, stacks of adjacent slices, or reliable anatomical landmarks.
Here, we propose a self-supervised 2D/3D registration approach to match a single 2D slice to the corresponding 3D volume. The method works in data without anatomical priors such as images of tumors. It addresses the dimensionality disparity and establishes correspondences between 2D in-plane and 3D out-of-plane rotation-equivariant features by using group equivariant CNNs. These rotation-equivariant features are extracted from the 2D query slice and aligned with their 3D counterparts.
Results demonstrate the robustness of the proposed slice-in-volume registration on the NSCLC-Radiomics CT and KIRBY21 MRI datasets, attaining an absolute median angle error of less than 2 degrees and a mean-matching feature accuracy of 89\% at a tolerance of 3 pixels.
\end{abstract}

Aligning a 2D object within a 3D volume poses a fundamental challenge across various disciplines, including medical imaging~\cite{chen2024optimization}, robotics~\cite{Gao2020-iz}, autonomous systems~\cite{DBLP:journals/corr/abs-2103-06432}, and augmented reality~\cite{WANG2020101890}. In medical imaging, the ability to conduct reliable 2D/3D registration enables combining information across multiple imaging modalities and dimensions for diagnosis, treatment decisions, therapeutic interventions or surgery~\cite{Markelj2010-kg,Nguyen2022-ed,chen2024optimization,Dong2023-cz,10.3389/frobt.2021.716007}. These existing approaches typically estimate the pose of a \textit{projection} image, such as an X-Ray, in relation to a 3D volume. They exploit the fact that the 2D image contains information of the entire volume. In this work, we propose a method to find the position of a 2D image slice in a 3D image volume. It is based on group rotation-equivariant features that enable the matching from 2D to 3D, even-though the image slice only contains a small fraction of the image volume information.

\section{Related work}
Current methods that address 2D/3D-registration are optimization-~\cite{DBLP:journals/corr/abs-1903-03896,Otake2013-ge}, learning-~\cite{Van_Houtte2022-ys,Dong2023-cz,xu2022svort}, or landmark-based \cite{10.3389/frobt.2021.716007}, or a combination of these techniques~\cite{Nath2019-fk}.
%Optimization part%
\paragraph{Optimization-based methods} are the most commonly used approach and iteratively adjust a 3D pose to match the similarity of the 2D data~\cite{DBLP:journals/corr/abs-1903-03896} starting from an initial pose estimate~\cite{10.3389/frobt.2021.716007}. Iterative approaches require a considerable number of optimization steps often entailing simulating 2D projections, leading to high computational costs~\cite{DBLP:journals/corr/abs-1903-03896,Nguyen2022-ed}, and are  sensitive to poor initialization~\cite{Nguyen2022-ed}.
%Key point landmark part%
\paragraph{Landmark and key-point based approaches} aim at reducing computational costs~\cite{evan2021keymorph} by relying on either  manual landmark annotations, or automated candidate point selection by approaches such as Canny edge detectors~\cite{Nguyen2022-ed}. Deep learning has made strides in automatically selecting landmarks with the assistance of labeled ground truths~\cite{Nath2019-fk}. 
Algorithms match key points either in 2D-to-2D~\cite{DBLP:journals/corr/abs-1712-07629,Lowe2004-if} or in 3D-to-3D~\cite{DBLP:journals/corr/abs-1902-05020,DBLP:journals/corr/abs-1809-05231}. However, it is still an open research question to accurately determine 2D key point positions in a 3D volume.
\paragraph{Deep learning-based 2D/3D-registration approaches} typically embed both the 2D slice and 3D volume into a shared feature space and directly predict the transformation parameters at once~\cite{Van_Houtte2022-ys,Dong2023-cz}. They enhance the conventional similarity measure by utilizing a neural network as their similarity function~\cite{chen2024optimization}. The biggest drawback of deep learning approaches is that they rely on large, paired datasets~\cite{Non-Rigid-2D-3D}. Moreover, the majority of learning-based methods require an optimization-based post-processing step to refine the result~\cite{chen2024optimization}. In fetal imaging learning based slice stack to volume registration approaches have been proposed for reconstruction~\cite{xu2022svort} and motion tracking~\cite{billot2023se3equivariant}. In~\cite{Casamitjana2022-aj} stacks of histology slices and brain MR data were registered for multi-modal high-resolution atlasing.

%Equivariant part and to get to our approach%
\paragraph{Equivariant representations} address the issue of learning based models often not exploiting the symmetries and equivariances present in the registration task.~\cite{evan2021keymorph} Recent methods address this issue by utilizing rotation-equivariance~\cite{wang2023robust,lin2023coarsetofine}. However, their applicability is confined to point clouds and consistent dimensionality such as 2D-to-2D~\cite{lee2023learning} or 3D-to-3D~\cite{billot2023se3equivariant,Equivariant-Filters}. For 2D/3D registration, existing methods are still limited by the need for manual initialization, a restricted search space or iterative refinement, and prior knowledge about the appearance and the geometry of the anatomical structures that are registered. This limits their utility in registering individual 2D slices of irregular structures such as tumors in 3D imaging data. 

\paragraph{2D projection to 3D volume}
A number of existing algorithms address the related task of registering a 2D X-ray image to a 3D CT volume. These algorithms use a 2D X-ray image, and generate a synthetic X-ray referred to as Digitally Reconstructed Radiograph (DRR)~\cite{staub2013digitally} using an assumed camera pose from the 3D volume. By comparing the synthetic X-ray and the provided image they optimize the camera pose to match the one corresponding to the observed 2D image. Similar to an X-ray, the DRR summarizes absorption along rays through the 3D volume. This approach is related to the problem addressed in this paper, but instead of estimating the pose of a camera yielding a given projection image, we estimate the pose of a single slice with arbitrary position and rotation within the 3D volume.

\paragraph{Contribution} Here, we propose \textit{SLIV-Reg}, a method for reliable 2D slice in 3D volume rigid registration. We use group-equivariant Convolutional Neural Networks (CNNs) to extract in-plane-rotation-equivariant features for candidate points in both slice and volume. To cope with out-of-plane rotations we efficiently sample 2D features in different orientations from the 3D volume. We then match a large number of corresponding 2D to 3D candidates, and predict the position of the 2D slice within the 3D volume using robust pose estimation. To reduce computational costs, we restrict candidate points to Canny edge detector points.
Compared to previous methods, we overcome the limitations of (1) manual landmark selection, (2) confining the search space,(3) pose initialization, (4) the need for iterative refinement steps, (5) the reliance on a stack of adjacent slices, (6) prior knowledge about the anatomical geometry of the imaged objects, and (7) the need of a 2D projection image. Our experiments demonstrate the robustness and accuracy of the proposed approach in CT data of lung tumors.

%%%%%%%%%%%%%%%%%%%%%%%%%%%%%%%%%%%%%
% METHOD SECTION
%%%%%%%%%%%%%%%%%%%%%%%%%%%%%%%%%%%%%
\section{Method}
\begin{figure}[th]
\includegraphics[width=\linewidth]{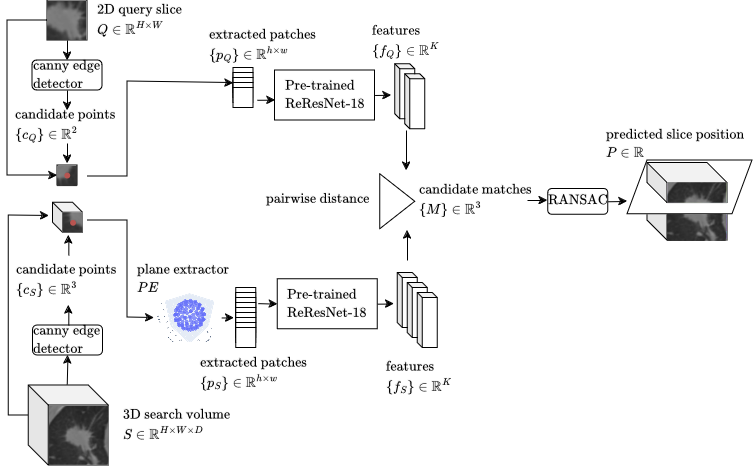}
\caption[Method overview]{SLIV-Reg model: a 2D query slice $\mathbf{Q}$ and a 3D search volume $\mathbf{S}$ are processed to identify 
candidate points \(\mathbf{c}_{Q}\) and \(\mathbf{c}_{S}\). At each point in $\mathbf{Q}$ one patch is extracted, and in $\mathbf{S}$ a set of R 2D patches in an equidistante sampling of orientations are extracted with the plane extractor (PE). A pre-trained ReResNet-18 encodes each patch to features \(\mathbf{f}_{Q}\) and \(\mathbf{f}_{S}\). For each \(\mathbf{f}_{Q}\) we find the closest feature vector \(\mathbf{f}_{S}\), resulting in a set of candidate matches $\mathbf{M}$. These matches serve as basis for a RANSAC estimate of the 2D slice pose in the 3D volume.}
\label{fig:grafical_abstract}
\end{figure}

We propose a method to achieve precise alignment between a 2D slice and a 3D volume through the utilization of rotation-equivariant feature matching (Fig.~\ref{fig:grafical_abstract}). For training, we conduct self-supervised pre-training of a rotation-equivariant encoder. During inference, we identify candidate points $\mathbf{c}_Q^i \in \mathbb{R}^2$, $i=1,\dots ,I$ on the 2D query slice $\mathbf{Q} \in \mathbb{R}^{H \times W}$ and $\mathbf{c}_S^j \in \mathbb{R}^3$, $j=1,\dots ,J$ in the search volume $\mathbf{S} \in \mathbb{R}^{H \times W \times D}$ with the help of Canny edge detector. For readability we omit the indices of candidate points through out the remaining manuscript, where not necessary. For the 2D query slice $\mathbf{Q}$ we extract patches $\mathbf{p}_Q$ at the candidate positions $\mathbf{c}_Q$. In the search volume $\mathbf{S}$ we extract a cube centered around $\mathbf{c}_S^j$ and extract $R$ patches $\mathbf{p}_S^{j,r}$ with different orientations with the Plane extractor (PE).
The pre-trained encoder is used to extract the feature vectors $\mathbf{f}_Q^i$ and $\mathbf{f}_S^{j,r}$ for all $\mathbf{p}_Q^i$ and $\mathbf{p}_S^{j,r}$. 
After computing the pairwise distances between $\mathbf{f}_Q$ and $\mathbf{f}_S$, the features with the lowest distance create pairs between $\mathbf{c}_Q$ and $\mathbf{c}_S$. These pairs are our candidate matches $\mathbf{M}\in \mathbb{R}^{3}$ and are utilized to robustly predict the best fitting position $P$ of the slice in the volume with the RANSAC algorithm. 

\paragraph{\textbf{Introduction to Steerable CNNs}}
Steerable CNNs implement E(2)-equivariant convolutions, meaning they are invariant to rotations and reflections on the image plane $\mathbb{R}^2$~\cite{e2cnn}. In these networks, the feature spaces consist of feature fields, which are characterized by a group representation that dictates how they transform in response to transformations of the input~\cite{e2cnn}.
\paragraph{\textbf{Rotation-equivariant pre-training}} 
We implement the rotation-equivariant ResNet-18~\cite{lee2023learning} as feature extractor, a model based on the ResNet-18 architecture but enhanced with rotation-equivariant convolutional layers from the E(2)-CNN framework~\cite{e2cnn} maps an image to a feature vector \(\mathbf{N}:\mathbf{p} \mapsto \mathbf{f}\). 
In the pre-training phase, we adopt the self-supervised setup employed by~\cite{lee2023learning}. The primary goal of pretraining is to enable the encoder to learn matching features in rotated images through extensive data augmentation~\cite{lee2023learning}. We trained on the non-small cell lung cancer (NSCLC)~\cite{Aerts2019-cy} dataset and on the Kirby21 dataset~\cite{landman2011multi}. This ensures that the extracted features are distinguishable not only by their spatial location but also by their in-plane rotation.

\paragraph{\textbf{Candidate points extraction in 2D and 3D data}} 
For matching sets of 2D to 3D points we first identify candidates at locations that have a chance to correspond. We assume that object edges are visible both in the query slice and the search volume, and use the Canny edge detector~\cite{4767851} to extract candidate points in the 2D- and 3D data. %This is facilitated by the presence of edges within a volume and the subsequently extracted slice. 
The query slice \(\mathbf{Q}\) undergoes blurring with a Gaussian kernel and is then processed through the Canny edge detector, revealing the positions of edge points \(\mathbf{c}_{Q}\). Similarly, we iterate over the z-axis of the search volume \(\mathbf{S}\) and find the candidate points \(\mathbf{c}_{S}\) accumulated over the volume slices.

\paragraph{\textbf{Patch extraction}} 
For each candidate point $\mathbf{c}_{Q}$ in the query slice $\mathbf{Q}$ we extract a patch \(\mathbf{p}_{Q}\in\mathbb{R}^{h\times w}\) centered around the point. 
For each candidate point $\mathbf{c}_{S}$ of the search volume we extract a cube centered around it. We then apply the \emph{plane extractor (PE)} to sample 2D planes out of this 3D sub-volume. Specifically, the cube rotates in R directions, guided by normal vectors on the unit sphere, and the central slice of the cube is extracted after the rotation. This leads to the extraction of R distinct patches  $\mathbf{p}_S^{j,r}\in\mathbb{R}^{h\times w}$ for R unique rotations, all centered around the candidate point $\mathbf{c}_{S}^{j}$.

\paragraph{\textbf{Rotation-equivariant feature extraction}} 
The pre-trained 2D ReResNet-18 encodes each patch $\mathbf{p}_Q$ and $\mathbf{p}_S$ to a feature vector $\mathbf{f}_{Q}=\mathbf{N}(\mathbf{p}_Q)\in\mathbb{R}^{K}$ and \(\mathbf{f}_{S}=\mathbf{N}(\mathbf{p}_S)\in\mathbb{R}^{K}\). In contrast to pre-training, we are not utilizing the group-alignment feature and multi-scale forward of~\cite{lee2023learning} during this step. 

\paragraph{\textbf{Feature matching}} 
For each query slice feature $\mathbf{f}_{Q}$ we identify the closest search volume feature $\mathbf{f}_{S}$ based on the L2 distance between feature vectors. 
Each 2D candidate point $\mathbf{c}_{Q}$ is matched to one 3D candidate point $\mathbf{c}_S$, resulting in a set of candidate matches $\mathbf{M}$.
\paragraph{\textbf{Slice position prediction}} 
To predict the final query slice position in the search volume, we use all candidate matches $M$ (i.e. corresponding $\mathbf{c}_S$ coordinates) as input to the RANSAC algorithm~\cite{10.1145/358669.358692}. RANSAC predicts the slice position $P = \langle \mathbf{n}, \mathbf{t}\rangle$ (normal vector, plane offset) that has the strongest support among the matched points. 

\begin{figure}[th]
\includegraphics[width=\linewidth]{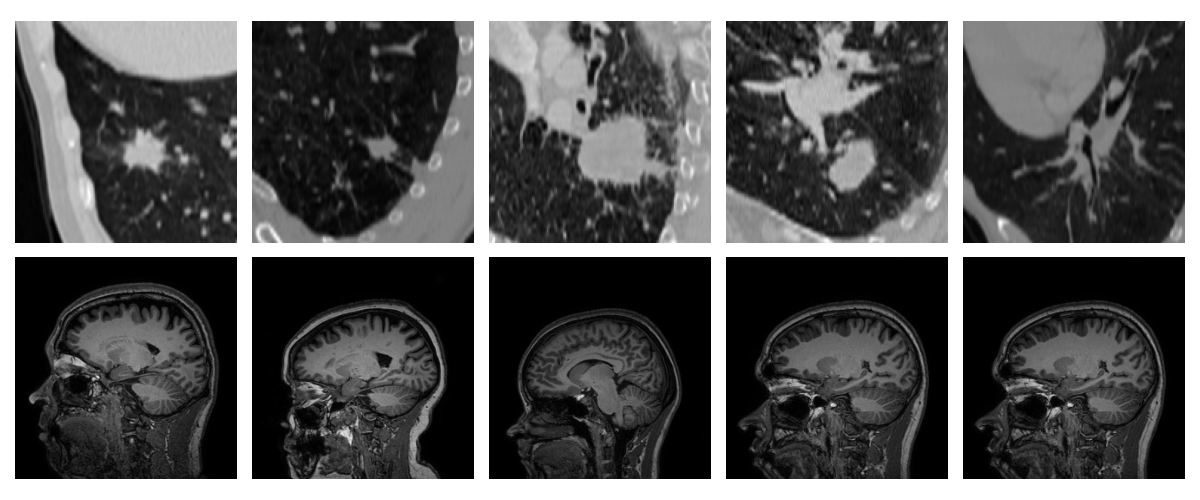}
\caption[blank]{Data used for method evaluation. First row: tumors in CT volumes of NSCLC patients illustrating the variability of tumor shapes; second row: T1-weighted MRI scans of the brain in the Kirby21 dataset.}
\label{fig:blank}
\end{figure}

%%%%%%%%%%%%%%%%%%%%%%%%%%%%%%%%%%%%%
% EXPERIMENTAL-SETUP SECTION
%%%%%%%%%%%%%%%%%%%%%%%%%%%%%%%%%%%%%
\section{Experimental Setup}
\textbf{Data:}
We used two medical imaging data sets of different modalities and anatomical structures for evaluation. The first dataset comprises 422 computed tomography (CT) volumes of subjects with non-small cell lung cancer (NSCLC) tumors~\cite{Aerts2019-cy}. We partitioned this data into training (335), validation (62), and test (25) sets. Each CT volume contains one tumor, and corresponding manual pixel-wise annotations. The second dataset - Kirby21~\cite{landman2011multi} - contains 41 T1 weighted magnetic resonance imaging (T1w MRI) volumes of the brain. For training and validation we used volumes KKI06 to KKI42,  for testing, we used volumes KKI01 to KKI05. Experiments A, B, and C are conducted using the NSCLC dataset, while Experiment D is performed using both datasets.

\textbf{Implementation details:}
The ReResNet-18~\cite{DBLP:journals/corr/abs-2103-07733} is pre-trained with 512x512 slices extracted from each volume. We follow the strategy of Lee~et~al.~\cite{lee2023learning} to enhance the model's geometric and photometric robustness by augmenting the slices with random affine homographies, gaussian and speckle noise addition, and motion blurring. We use a cyclic group order of 16, a batch size of 16, a learning rate of 10-4, weight decay of 0.1 and 12 epochs with 322 iterations. The extracted patches $p_Q$ and $p_S$ at the location of the candidate points $\mathbf{c}_Q$ and $\mathbf{c}_S$ have a size of 21$\times$21. 
Features $\mathbf{f}_Q$ and $\mathbf{f}_S$ have a feature size $K$ of 512. The parameter of the RANSAC~\cite{10.1145/358669.358692}~\footnote{https://github.com/leomariga/pyRANSAC-3D} algorithm are a treshold of 1 pixel, which defines the distance to be counted as a inlier, maximum iterations equals 1000, and minPoints was set to 40.
The pretraining for the Rigid Transformation Parameter Initialization (RTPI) network is done based on the RTPInetV3~\footnote{https://github.com/m1nhengChen/SOPI} as the model~\cite{chen2024embedded} with a change of the PRO-ST~\footnote{https://github.com/gaocong13/Projective-Spatial-Transformers} module~\cite{gao2020generalizing}, to the standard pytorch grid sampler to extract the centre slice instead of the resulting 2D projective image. We use a batch size of 16, a learning rate of 10-3, stepLR scheduler with gamma of 0.5 every 30 epochs with 100 epochs in total.
Pretraining has been conducted separately on both datasets using the same setup and configurations.
For the CMA-ES~\footnote{https://github.com/m1nhengChen/CMAES-reg} there is no pretraining needed. The hyperparameters specify that the initial rotation matrix should not include any rotation of the volume, a sigma of 10-3, no lr adaptation and population size.
For the experiments on the NSCLC dataset, subvolumes are specified to do the inference on the tumor tissue. For both datasets, the volume is cropped after rotation to eliminate any indications of rotation, such as existing black borders.
All experiments were conducted on a machine with an AMD Ryzen 5 3600 CPU and a NVIDIA GeForce RTX 2060 Super. 

\paragraph{\textbf{Experiment A: Accuracy of 2D slice in 3D volume pose estimation}}
To evaluate the registration accuracy of the proposed algorithm, we randomly sample locations and orientations of 2D slices in a 3D volume, and test whether the model can accurately estimate the pose based on the 2D query slice, and a 3D volume.  For each 3D volume of the test set, we extract 90 2D slices intersecting the bounding box of the tumor. First, 30 slices are sampled in 30 random approximately equidistant directions. Then, slices with the same orientations, but a random translation off-set and distance of -6 and +6 voxels are extracted resulting in overall 90 slices varying in orientation and location. We use the proposed algorithm to perform registration, estimating the pose of each 2D query slice in the corresponding 3D search volume. Registration accuracy is evaluated by comparing the angle difference of the orthogonal vectors of predicted and the ground-truth slice position. % and (2) computing the mean matching accuracy (MMA) with pixel thresholds of 3/5/10 pixels. 
We vary the number R of directions used be the plane extractor from 10 to 700 to assess the influence of this hyper parameter on accuracy.

\paragraph{\textbf{Experiment B: Accuracy of 2D/3D image feature matching}}
In addition to the accuracy of the estimated pose, we evaluate the mean matching accuracy (MMA) of feature points in the correct position of the 2D query slice, and its estimated position with pixel thresholds of 3/5/10 before and after RANSAC matching~\cite{10.1145/358669.358692}. MMA evaluates the fraction of correctly matched feature points being located within a certain radius or "pixel threshold".

\paragraph{\textbf{Experiment C: Sensitivity of the feature distance to angle differences}}
The model yields a feature distance between points in the 2D query slice and slices extracted from the 3D search volume. To facilitate correct point matching, the feature distance should be relatively small for the same point and different angles, and large for different points in the 3D volume. To investigate this behavior, we compare the feature distance, when gradually changing orientation (intra-point feature distance) with the feature distance between slices sampled at different positions (inter-point feature distance).
For each candidate point, we extract 400 features $\mathbf{f}_S$ resulting from different orientations ($R=400$). We then sample pairs of random orientations at the same point (reference feature A, intra-point feature B) with an angle difference 10 to 90 degree and compute their feature distance ($\beta$). We then sample a feature with the same angle difference but at a different random candidate point (inter-point feature C) and compute the feature distance to A ($\gamma$). $\beta$ should increase with increasing angle difference, $\beta$ should be lower compared to $\gamma$, and the difference $\beta - \gamma$ should decrease with increasing angle.

Additionally, we extract 360 in-plane features $\mathbf{f}_S$ for each candidate point in the volume, by conducting the plane extraction only within the xz-plane using 360 equidistant directions. We use the feature at the 0-degree angle as reference and calculate all intra-and inter-point feature distances. We evaluate the first and second property by plotting the intra-point feature distances against the angle, and comparing intra- and inter-point feature distances, respectively. 
The metric used to assess the distance between features is the L2 norm.

\paragraph{\textbf{Experiment D: Comparison with state-of-the-art methods}}
To our knowledge, there is no prior algorithm addressing the exact same problem. However, 2D projection image to 3D volume registration approaches exist, and the comparison is instructive. We compare the proposed algorithm with two existing approaches. First, we apply an optimization based algorithm called Covariance matrix adaptation evolution strategy (CMA-ES) as proposed by~\cite{chen2024optimization}, second, we apply the Rigid Transformation Parameter Initialization (RTPI) module suggested by~\cite{chen2024embedded}. We evaluate both algorithms for the median absolute angle error without any translation on 30 slices against SLIV-Reg with $R=300$ and $R=700$ directions on the sphere.

%%%%%%%%%%%%%%%%%%%%%%%%%%%%%%%%%%%%%
% RESULTS SECTION
%%%%%%%%%%%%%%%%%%%%%%%%%%%%%%%%%%%%%
\section{Results}
\begin{figure}[t]
\includegraphics[width=\linewidth]{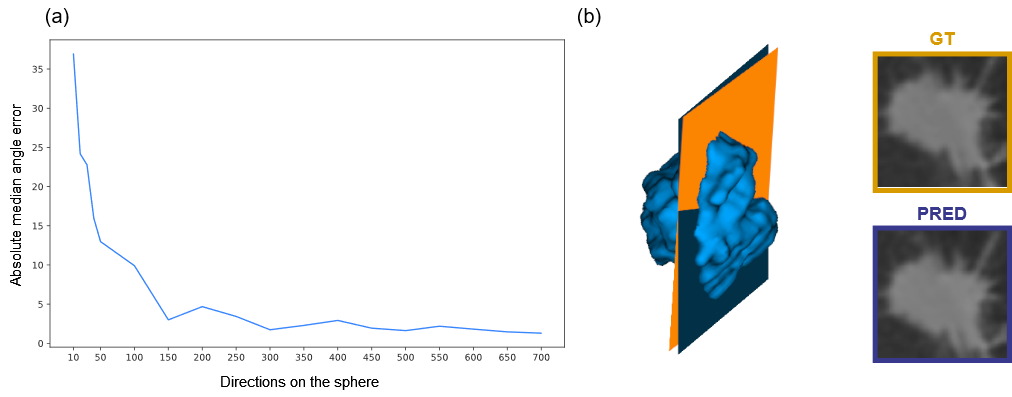}
%\minipage{0.50\textwidth}
%  \includegraphics[width=\linewidth]{figures/slice_angle_error.png}
%\endminipage\hfill
%\minipage{0.50\textwidth}
%  \includegraphics[width=\linewidth]{figures/3D_with_matched_slices_new.png}

%\endminipage\hfill
\caption[experimentA]{Experiment A: (a) The median absolute angle error for varying directional sampling density of the plane extractor. (b) Qualitative 3D visualization of a registration result, showing the predicted and ground truth slice pose, together with their 2D projections on the right hand side.}
\label{fig:angle_error_qualitative}
\end{figure}

%Experiment A:
Fig.~\ref{fig:angle_error_qualitative}(a) shows the median absolute angle error between true and estimated post of the query slice as a function of directional sampling density $R$. SLIV-Reg yields a median absolute error of $\le$ $3\textdegree$ for  $R\geq150$, and the error converges after that. Fig.~\ref{fig:angle_error_qualitative}(b) provides an example visualization of the estimated and true pose of the query slice, and the corresponding images extracted from the 3D volume (GT at the correct pose, PRED at the estimated pose). The example illustrates the level of similarity after good matching. 

\begin{figure}[t]
\includegraphics[width=\linewidth]{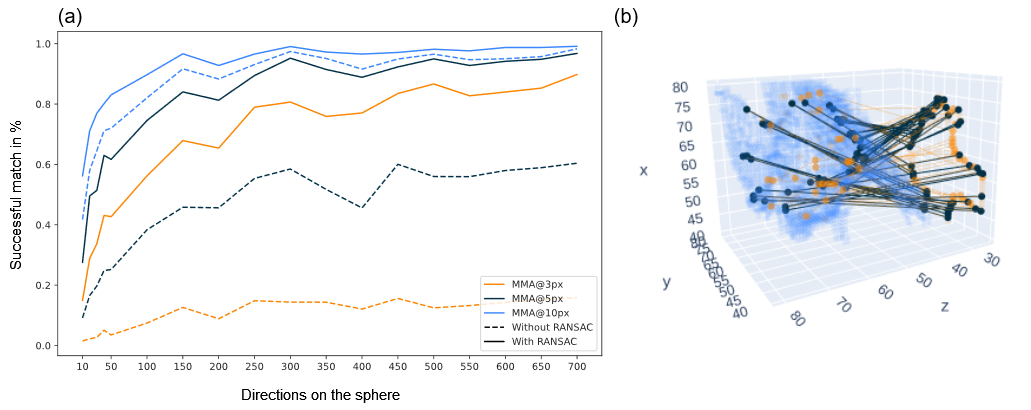}

\caption[experimentB]{Experiment B: (a) The MMA with pixel thresholds 3/5/10 before (dashed line) and after RANSAC (solid line) for varying directional sampling densities R. (b) 3D visualization of a matching result: $\mathbf{c}_S$ candidate points are shown in blue in the 3D space, $\mathbf{c}_Q$ points are visualized in the yz-plane on the right hand side, and  accurate/inaccurate matches are highlighted in green/red (MMA pixel threshold of 3).}
\label{fig:success_mma}
\end{figure}

%Experiment B
Fig.~\ref{fig:success_mma}(a)  shows the 2D/3D feature matching evaluation results as MMA as a function of directional sampling density of the plane extractor. Starting from $R\geq300$, $\ge$80\% of feature points are successfully matched (@3px) after RANSAC. This is also reflected in qualitative results illustrated in Fig.~\ref{fig:success_mma}(b), where the candidate points of the query slice on the right are connected to their corresponding matches among all 3D candidate points, highlighted in blue. Results show a modest enhancement of 1.64\% from 97.42\% to 99.06\% before and after RANSAC at an error rate of 10 (MMA@10px) and $R=300$. A notable improvement before and after RANSAC is observed for MMA@3px, with a substantial increase of 66.26\%, rising from 14.39\% to 80.664\% for $R=200$. 

\begin{figure}[t]
\includegraphics[width=\linewidth]{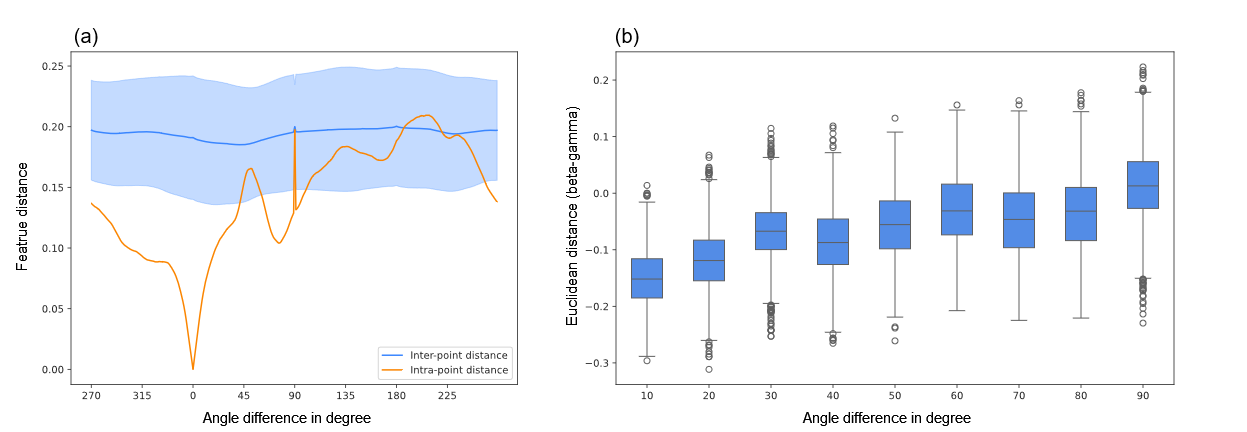}

\caption[Intra/inter-point distance analysis]{Experiment C: (a) For varying angles, intra-point feature distances are illustrated in orange, while inter-point distances are highlighted in blue, accompanied by standard deviation. (b) Median difference between intra- and inter-point feature distances at given directions, shown for absolute angle differences of 10 to 90 degrees.}\label{fig:360-intra-inter}
\end{figure}

%Experiment C
Fig.~\ref{fig:360-intra-inter}(a) shows the feature distance for slices extracted at the same location with different orientations (orange) and slices extracted at different locations and the corresponding orientations (blue) in an interval of one degree. Intra-point feature distances are markedly below the feature distances to other points up to an angle difference of appr. $\pm40\textdegree$. This is consistent with the observation that a sampling density of 150 directions on the sphere achieves a median angle error of $\le$ 5 degrees. The sampling directions have an approximately $8\textdegree$ offset, which is within the acceptable range. In line with these results, the median difference between the intra- and inter-point feature distances ranges from -0.15 to 0.01, with increasing difference for decreasing angles (Fig.~\ref{fig:360-intra-inter}(b)).

\begin{figure}[th]
\begin{center}
\includegraphics[width=.7\linewidth]{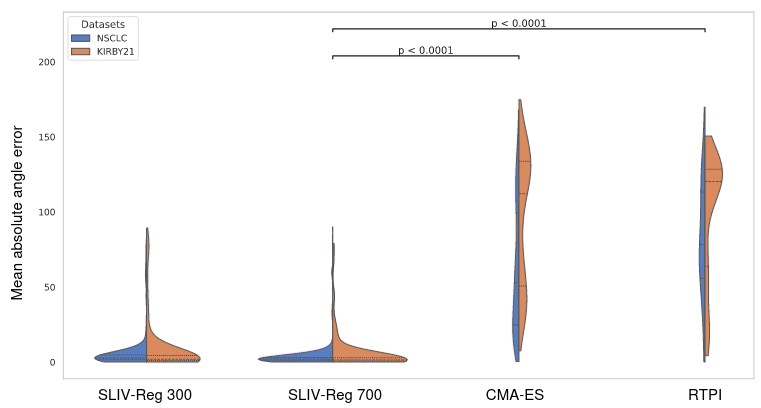}
\end{center}
\caption[experimentD]{Experiment D: Median absolute angle error between the ground truth slice and the predicted slice. Each method was evaluated using both the NSCLC and Kirby21 datasets. P-values indicate the statistical significance of our method.}
\label{fig:stateoftheart}
\end{figure}

The quantitative comparison of SLIV-reg with two state-of-the-art approaches is shown in Fig.~\ref{fig:stateoftheart}. We evaluated two sampling density values for SLIV-reg ($R=300$ and $R=700$) with CMA-ES and RTPI. Experiments were performed on two different data sets (CT tumor, MRI brain). We compare the mean absolute angle error of the estimated query slice pose, and the correct pose. Results demonstrate a median absolute angle error of $\le$ 2 degree on both NSCLC and Kirby21 datasets for the proposed SLIV-reg with $R=700$. The CMA-ES algorithm depends heavily on its starting point. It can find the correct position with an angle error of $\le$ 0.5 degree if the initial rotation matrix is close to the true one. However, if the initial matrix is substantially different to the correct orientation, the algorithm cannot converge as seen in Fig.~\ref{fig:stateoftheart} where initialization was chosen randomly. The Rigid Transformation Parameter Initialization Module (RTPI) was trained unsuccessful and could not converge. We believe the reason is the same as for the CMA-ES method. Both methods rely on predicting a rotation matrix from the projection image and volume. This is feasible because the projection image contains more information about the entire image, compared to the relatively minimal information of an extracted slice. 

\begin{table}[t]
  \caption{Absolute angle error between the ground truth slice and predicted slice of our method against the baselines on the NSCLC and Kirby21 datasets.}
  \label{sample-table}
  \centering
  \begin{tabular}{lcccccc}
    \toprule
    Method & \multicolumn{3}{c}{NSCLC} & \multicolumn{3}{c}{Kirby21} \\
    \cmidrule(r){2-4} \cmidrule(r){5-7}
    & Mean & Median & STD & Mean & Median & STD \\
    \midrule
    RTPI         & 96.43 & 112.07 & 45.34 & 99.05 & 120.26 & 42.46 \\
    CMA-ES       & 61.97 & 52.81  & 43.36 & 82.14 & 78.29  & 39.88 \\
    SLIV-Reg 300 &  6.37 &  2.71  & 13.48 &  2.85 &  1.23  &  8.83 \\
    SLIV-Reg 700 &  4.01 &  1.93  &  9.63 &  5.87 &  1.33  & 13.80 \\
    \bottomrule
  \end{tabular}
\end{table}

%%%%%%%%%%%%%%%%%%%%%%%%%%%%%%%%%%%%%
% DISCUSSION & CONCLUSION SECTION
%%%%%%%%%%%%%%%%%%%%%%%%%%%%%%%%%%%%%
\section{Discussion and Conclusion}
We propose \textit{SLIV-Reg}, a slice-in-volume registration approach that reliably predicts the pose of a 2D query slice in a 3D search volume based on rotation-equivariant 2D-3D feature matching and robust pose estimation. Results on CT data of lung tumors and MRI data of the brain show that it can accurately identify the location and rotation of a slice, independent of prior knowledge of anatomy. This makes it useful for matching imaging data of irregular structures such as tumors. In contrast to existing algorithms, the method does not require initialization, iterative refinement, constraints of the search space, stacks of neighboring slices, or an image that is a projection of the 3D volume.
Our method is inspired by recent 2D-to-2D registration approaches~\cite{lee2023learning} that attain an MMA@3px of 91.69\% for the Roto-360 dataset. We introduce a 2D-to-3D matching approach that can cope with in-plane and out-of-plane rotations. It achieves comparable quantitative results in a 2D-to-3D task, reporting a MMA@3px of 89.82\% for the NSCLC-Radiomics dataset. This indicates that SLIV-Reg successfully addresses the challenge of appropriately matching features across 2D and 3D domains for rigid registration by sampling multiple rotation-equivariant 2D features for a single 3D position. 
Accuracy benefits from increased directional sampling density R of the 3D image descriptors, but makes matching feasible already with low directional sampling density. This is due to the feature distances being relatively insensitive to orientation changes compared to features extracted at different positions (Fig.\,\ref{fig:360-intra-inter}a). 

The lack of 2D-in-3D registration baselines leads us to compare our algorithm with the similar problem of X-ray to CT volume registration. We believe this works because the projection DDR image has minimal information loss due to its summative nature, providing more information about its exact location in the volume. In our problem setting, only a small part of the volume is present in the 2D slice. For example, a subvolume of a tumor in a lung CT scan of size (100,100,100) with a 2D slice size of (100,100) results in an information loss of 99.9\%. In our comparative evaluation, this leads to incorrect registrations using the baseline methods for our problem as they are sensitive to good initialization. As shown in Fig.~\ref{fig:stateoftheart}, CMA-ES can converge to a prediction within 1 degree if the initialization is close.

\section{Limitations and Future Work} %"The authors are encouraged to create a separate "Limitations" section in their paper"
While future work will evaluate translation to different modalities and non medical datasets,  rotation-equivariant networks have shown to generalize well with little risk of overfitting~\cite{e2cnn,Bylander1779131}. As no other comparable 2D-3D single slice registration approach has been proposed, we rely on the comparable 2D-2D registration approach~\cite{lee2023learning} and similar learning tasks as 2D X-ray to 3D CT volume as reference when comparing to the state of the art. 

The proposed method offers a means to perform 2D-3D slice-in-volume registration accurately and reliably. Comparably to existing 3D-3D machine learning based approaches it offers the perspective of integrating multi-modal data such as in-vivo and ex-vivo imaging data.
%Hinzugefügt wegen der Appendix Frage 10
A wider positive impact on society can be achieved by enhancing the alignment of histology images with CTA scans and improving the accuracy of predicting treatment responses.

\begin{credits}
\subsubsection{\ackname}
The financial support by the Austrian Federal Ministry for Digital and Economic Affairs, the National Foundation for Research, Technology and Development and the Christian Doppler Research Association, Siemens Healthineers, the Austrian Science Fund (FWF, P 35189-B - ONSET), and the Vienna Science and Technology Fund (WWTF, PREDICTOME [10.47379/LS20065]) is gratefully acknowledged.

\subsubsection{\discintname}
The authors have no competing interests to declare that are relevant to the content of this article.

\end{credits}

\bibliographystyle{splncs04}

\end{document}